# LOOKING AND LISTENING INSIDE AND OUTSIDE: MULTIMODAL ARTIFICIAL INTELLIGENCE SYSTEMS FOR DRIVER SAFETY ASSESSMENT AND INTELLIGENT VEHICLE DECISION-MAKING

**Ross Greer[1,2], Laura Fleig[1,2,3], Maitrayee Keskar[1, 2], Erika Maquiling[1], Giovanni Tapia Lopez[1], Angel Martinez-Sanchez[1], Parthib Roy[1], Jake Rattigan[4], Mira Sur[4], Alejandra Vidrio[4], Thomas Marcotte[4], and Mohan Trivedi[2]**

[1] Machine Intelligence, Interaction, and Imagination (Mi$^3$) Laboratory, University of California, Merced, USA
[2] Laboratory for Intelligent and Safe Automobiles (LISA), University of California, San Diego, USA
[3] Johns Hopkins University, USA
[4] Center for Medicinal Cannabis Research (CMCR), University of California, San Diego, USA

## ABSTRACT

The looking-in-looking-out (LILO) framework has led to intelligent vehicle applications which understand both the outside scene and the driver state to lead to improved safety outcomes, with examples in smart airbag deployment, takeover time prediction in autonomous control transitions, driver attention monitoring, and more. In this research, we propose an augmentation to this framework, making a case for the audio modality as an additional piece of important information to understand the driver—and, in the evolving autonomy landscape, also the passengers and those outside the vehicle. We expand the LILO framework by incorporating audio signals, forming the looking-and-listening inside-and-outside (L-LIO) framework to enhance driver state assessment and environment understanding through multimodal sensor fusion. We evaluate three example cases where audio enhances vehicle safety: supervised learning on an audio dataset of driver speech to classify potential impairment states (e.g., intoxication), collection and analysis of natural language instructions from passengers (e.g., "turn after that red building") to motivate how spoken language can interface with planning systems through audio-aligned instruction data, and shortcomings in vision-only systems which may use audio as a disambiguating cue to understand the guidance and gestures of external scene agents. Datasets include custom-collected in-vehicle and external audio samples in both controlled and real-world environments. We share pilot findings within these three case examples, showing that the inclusion of audio yields safety-relevant insights, particularly for nuanced or context-rich scenarios where sound is critical to safe decision-making or visual signals alone are insufficient. Challenges include ambient noise interference, privacy considerations, and robustness across human subjects, motivating further work to evaluate system reliability in dynamic real-world contexts. L-LIO augments driver and scene understanding through multimodal fusion of audio and visual sensing, offering new paths for safety intervention.

## INTRODUCTION

Ensuring safety in increasingly automated vehicles requires a comprehensive understanding of both the external driving environment and the internal state of vehicle occupants, motivating the development of looking-in, looking-out (LILO) framework [1], which integrates outward-facing perception of the roadway with inward-facing sensing of driver behavior and state. LILO-style systems have enabled advances in driver monitoring, attention estimation, takeover readiness prediction, and safety-critical interventions such as adaptive warnings and smart restraint deployment [12-13, 40-41]. These approaches have largely relied on visual sensing—cameras observing the road, the driver, and the vehicle interior—often augmented with vehicle telemetry and inertial measurements.

However, as vehicles become more intelligent, interactive, and socially embedded, vision alone is insufficient to fully capture the contextual cues that influence driving safety. Human drivers routinely rely on audio information—speech, tone, alarms, sirens, horns, shouts, and other acoustic signals—to assess risk, infer intent, and adapt behavior [26, 42-45]. These signals are often temporally urgent, semantically rich, and difficult or impossible to infer reliably from visual data alone. Despite their importance, audio signals are rarely included in contemporary large-scale autonomous driving and driver monitoring datasets, which predominantly emphasize visual and LiDAR-based perception [46-50]. This omission limits the ability of learning-based systems to reason about safety-critical situations that are inherently multimodal. Moreover, recent advances in speech processing, foundation audio models, and multimodal representation learning make it timely to revisit the role of audio as a scalable, learnable input for intelligent vehicle systems.

Beyond its role as a perceptual cue, audio has taken on renewed importance with the emergence of large language models (LLMs), vision-language models (VLMs), and vision-language-action (VLA) systems, which increasingly rely on natural language as a central interface for reasoning, instruction following, and decision-making [33-35, 51-58]. In this new paradigm, spoken language is not merely an auxiliary signal but a primary carrier of semantics, intent, and task specification. For intelligent vehicles, this shift fundamentally changes how driving information can be represented, learned, and acted upon.

Spoken audio provides a natural mechanism for attaching linguistic structure to dynamic driving scenes. Verbal instructions ("turn after that red building," "slow down near the cones"), descriptions ("there's a pedestrian stepping out"), and contextual alerts ("sirens approaching from behind") directly link perception to high-level reasoning in a way that is well aligned with modern foundation models. Unlike pre-defined symbolic interfaces or hand-engineered command grammars, language-based interaction enables flexible, compositional, and context-aware communication between humans and vehicles—capabilities that are increasingly expected in semi-autonomous and autonomous driving systems.

At the same time, recent advances in multimodal foundation models have demonstrated strong generalization when trained on paired vision-language or audio-language data, but these models are rarely grounded in real-world driving contexts [56, 59]. Without explicit support for audio-based language grounding both inside and outside the vehicle, intelligent transportation systems risk being disconnected from the very modalities that underpin state-of-the-art AI reasoning. Incorporating audio into vehicle sensing therefore serves a dual purpose: it enhances safety through additional perceptual channels, and it enables vehicles to participate fully in the language-centric AI ecosystem that is rapidly shaping robotics and autonomous systems research.

Motivated by these trends, we introduce the looking-and-listening-inside-and-outside (L-LIO) framework as an extension of LILO that explicitly elevates audio to a first-class modality for safe autonomous driving, incorporating audio signals from both the vehicle interior and the surrounding environment and acknowledging the changing source of the 'driver' from the human touching the steering wheel to a combination of human agents and automated systems within and outside the vehicle. L-LIO treats audio not merely as an auxiliary cue, but as a complementary modality that can directly inform driver state assessment, human-vehicle interaction, and environment understanding. This perspective reflects both the realities of human driving behavior and the evolving role of intelligent vehicles as interactive agents operating in complex, multimodal social environments. L-LIO is not simply a sensor-level augmentation; rather, it is a systems-level abstraction that connects multimodal sensing with modern AI architectures capable of reasoning over language, perception, and action. By jointly modeling visual and auditory signals from both the vehicle interior and exterior—beyond its inclusion of audio as an informative cue to driver and surrounding states—L-LIO provides a principled foundation for integrating LLMs, VLMs, and VLA-style planners into safety-critical vehicle decision-making loops.

Within this framework, audio supports multiple complementary roles: as a diagnostic signal for driver state, as a natural interface for human instruction and intent expression, and as an external environmental channel conveying urgent safety information. Crucially, these roles map directly onto the strengths of contemporary multimodal learning systems, enabling richer representations of driving context than vision-only pipelines can provide.

In this paper, we operationalize the L-LIO framework through three case studies that reflect these roles. We examine (i) driver speech analysis for assessing impairment and correlating predictions with driving performance metrics, (ii) passenger-spoken instruction parsing for grounding natural language in scene-aware planning and dataset creation to enable these abilities at scale, and (iii) limitations of vision to identify safety-critical alerts and socially meaningful signals in the driving environment, motivating the need for acoustic event detection and classification. Together, these studies demonstrate how listening, alongside looking, allows intelligent vehicles to leverage recent advances in multimodal artificial intelligence while addressing concrete safety challenges.

L-LIO augments the LILO framing towards automated driving paradigms, where the concept of vehicle 'driver' morphs in multiple directions. While LILO integrates the driver as a (presumed human) agent in the natural driving control flow the primary human of relevance (both in- and out-of-cabin), L-LIO considers the driver an optional or intermittent driving agent, where the vehicle itself may be the driver in the automated context, and now elevates the system's observations and predictions of all humans both inside and outside such that their intentions and instructions become critical context to safe automated driving. While LILO frames the driver as the interface between the driving environment and the vehicle, L-LIO considers the vehicle to be directly connected in the driving flow, with the driving system drawing information from the human and the environment to operate safely when automated, and still affording for the "natural driving flow" of LILO in cases of driving control transitions from automated to manual [40]. The prevalence of language as a mechanism for understanding these intentions and instructions elevates audio from a "looking" framework to a "looking and listening" framework.

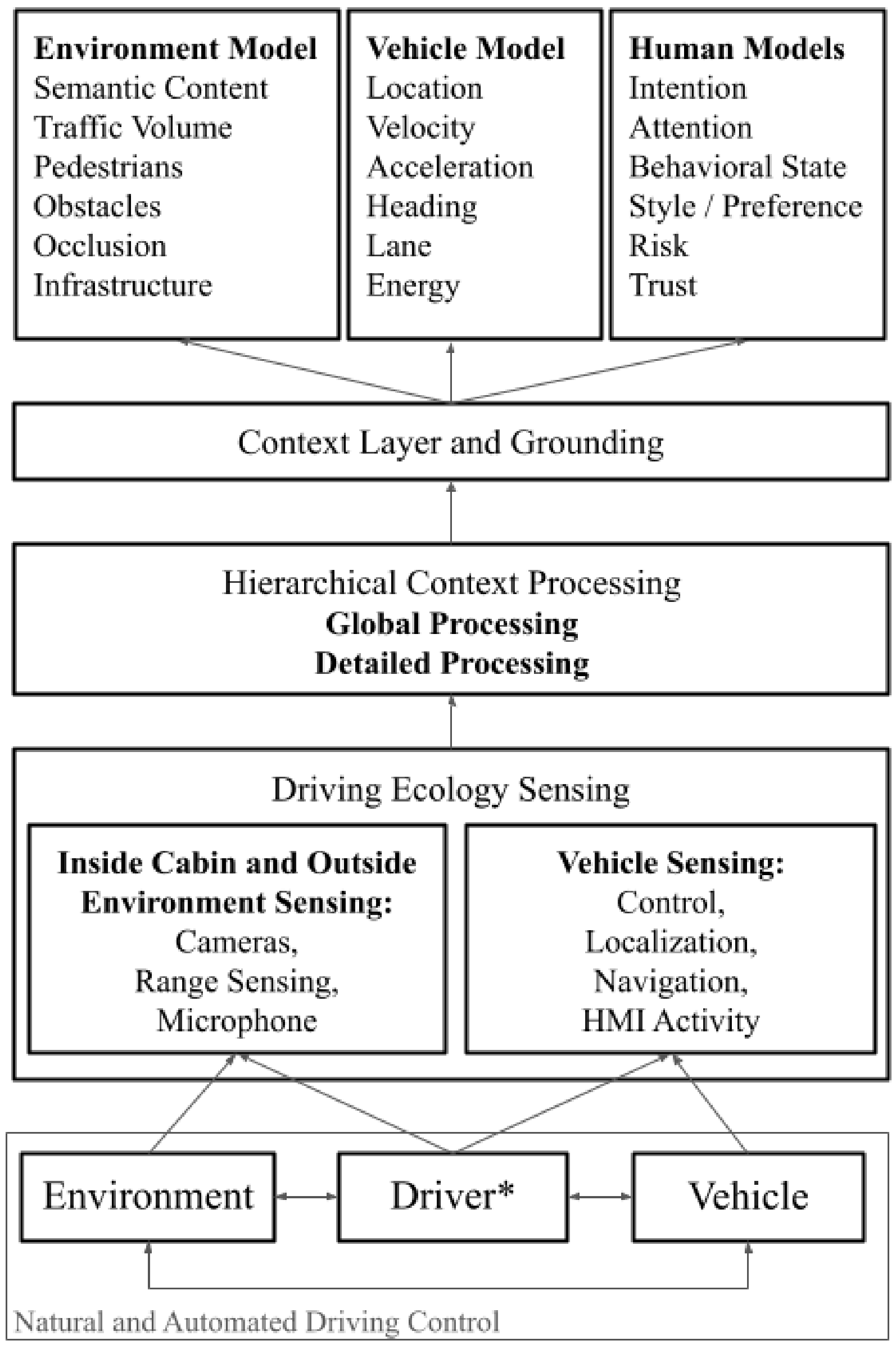

***Figure 1. L-LIO framework diagram, augmenting the LILO framework [1] in automated driving safety context. We refer to the Driver as Driver* in our framing, signifying that this vehicle occupant may intermittently control the vehicle, or, may be an occupant providing driving instructions or observable information to an automated driving system.***

By framing these contributions within a unified L-LIO perspective, this work aims to motivate broader inclusion of audio in intelligent vehicle datasets and systems, and to highlight its role in next-generation multimodal safety assessment for next-generation intelligent vehicles. Rather than proposing a single monolithic model, we present evidence across multiple tasks and data sources that collectively support the inclusion of audio as a core component of multimodal vehicle intelligence that is essential for scalable, interactive safety in the era of large multimodal AI models and increased vehicle autonomy.

## RELATED RESEARCH

### Looking-In, Looking-Out and Driver Monitoring Systems

The looking-in, looking-out (LILO) paradigm has become a foundational concept in intelligent vehicle safety systems, emphasizing the joint modeling of the external driving scene and the internal state of the driver. Early work demonstrated that driver monitoring—particularly gaze, head pose, and attention estimation—plays a critical role in safety-critical scenarios such as automation handover and driver distraction mitigation. Trivedi et al. articulated the importance of integrating inward- and outward-facing vision systems for situation awareness and safety assessment in intelligent vehicles, establishing a basis for multimodal driver-centric perception systems [1].

Subsequent research expanded LILO-style systems to include driver pose estimation, gaze tracking, and cognitive state inference using in-cabin cameras and vehicle telemetry [6, 7, 8, 39]. Comprehensive surveys have documented advances in vision-based driver monitoring, including drowsiness detection, distraction analysis, and workload estimation [9, 10, 11]. These systems have enabled applications such as adaptive driver assistance, takeover readiness estimation, and personalized warnings [12, 13]. However, the majority of LILO systems remain heavily vision-centric, with limited exploration of naturalistic human communication modalities such as speech and environmental sound.

### Driver State and Impairment Detection

Detecting unsafe driver states, including fatigue, distraction, and intoxication, has been an active research area for decades [16-18]. Vision-based indicators such as eyelid closure, gaze dispersion, and head motion have been widely used, alongside vehicle-based measures such as lane deviation, steering variability, and reaction time [19, 20, 16]. While effective, these indicators can be indirect and influenced by confounding factors such as road geometry, traffic conditions, and individual driving style.

Speech and audio analysis has been explored as a complementary modality for assessing cognitive and physiological state in domains such as healthcare, human–computer interaction, and affective computing. Prior work has shown that speech characteristics can reveal fatigue, stress, and intoxication through changes in prosody, articulation, and spectral features [21-24]. In vehicular contexts, however, speech-based impairment detection remains relatively underexplored and is rarely evaluated alongside driving performance metrics. This gap motivates further investigation into audio as an early and semantically rich signal for driver state assessment within safety-critical systems.

### Audio and Acoustic Event Recognition for Transportation Safety

Audio perception plays a crucial role in human driving, particularly for detecting emergency vehicles, horns, and other road users outside the visual field of view. Early transportation research explored acoustic siren detection and localization for emergency vehicle awareness using signal processing techniques [25, 26]. More recent work has applied machine learning methods to environmental sound classification, including urban sound and traffic noise recognition [27-29].

Despite these advances, modern autonomous driving datasets and perception stacks overwhelmingly prioritize visual, LiDAR, and radar sensing. As a result, safety-critical acoustic events—such as occluded sirens or shouted human warnings—are largely absent from learning-based vehicle perception pipelines. In contrast, recent work in embodied AI and robotics has begun to revisit sound as an informative modality for navigation and interaction, highlighting its potential value for safety-aware decision-making.

**Natural Language Interaction and Language-Grounded Driving**
Natural language has emerged as an increasingly important interface for human-vehicle interaction, particularly in advanced driver assistance and autonomous navigation systems. Early in-vehicle dialogue systems focused on constrained command grammars for navigation and infotainment control [30]. More recently, research in robotics and embodied AI has demonstrated that free-form language instructions can be grounded in perception to guide navigation and action [31, 32].

Inspired by these advances, autonomous driving research has begun exploring language-conditioned perception, planning, and scene understanding. Vision–language models have been applied to tasks such as driving scene description, instruction following, and question answering [33-35]. However, much of this work relies on text-based inputs or offline annotations, rather than real-time spoken language grounded in multimodal vehicle sensing. The role of audio as both a perceptual signal and a linguistic interface remains underdeveloped in this literature.

**Multimodal Foundation Models and Vehicle Intelligence**
Recent progress in large-scale multimodal foundation models, including vision–language and language–action architectures, has reshaped research in robotics and embodied intelligence. Models such as CLIP demonstrated the power of learning joint visual–linguistic representations from large-scale paired data [36], while subsequent work extended these ideas to action-conditioned and instruction-following systems. These models enable flexible reasoning, generalization, and compositional task specification through natural language.

In autonomous driving, early efforts have begun exploring foundation-model-based representations for perception and planning [37, 38]. However, most vehicle-focused applications remain constrained by the modalities available at inference time, often excluding audio or treating language as an offline annotation rather than a live signal. Without explicit support for listening inside and outside the vehicle, intelligent driving systems cannot fully leverage advances in language-centric AI.

In contrast to prior work, this paper frames audio not merely as an auxiliary cue, but as a first-class modality that supports driver state assessment, language-grounded interaction, and external environment understanding. By extending LILO to the looking-and-listening inside-and-outside (L-LIO) framework, we provide a unifying perspective that connects multimodal vehicle sensing with modern language- and action-centric AI models. Our case studies complement existing vision-centric approaches while addressing a critical gap in current datasets and systems, motivating broader inclusion of audio in next-generation intelligent vehicle safety research.

## METHODS
The relevance of audio has increased alongside several converging trends. First, driver state assessment increasingly seeks to identify subtle or transient impairments—such as intoxication, fatigue, or emotional distress—that may manifest through speech characteristics before they are evident in driving kinematics or visual behavior. Second, natural language interaction between passengers and vehicles is becoming commonplace, particularly in partially and fully automated driving, where spoken instructions such as “turn after that red building” or “pull over up there” must be grounded in the perceived scene to guide planning and control. Third, external acoustic cues—including emergency sirens, car horns, human shouting, or traffic director whistles—often carry immediate safety implications and require rapid behavioral adaptation, even when the visual source of the signal is occluded or ambiguous.

Accordingly, we structure the remainder of this paper within three case studies examining the role of audio towards in-cabin perception, grounding of in-cabin language to the external scene, and dynamic scene understanding. These three non-exhaustive cases present early examples where an L-LIO sensing framework expands capabilities in perception and decision-making for intelligent vehicles, making increased safety possible by altering driver behaviors and navigating complex, risky traffic scenarios.

As detailed in the following sections on methodology and results, our case studies are supported by real-world datasets collected from observing human participants of naturalistic driving simulations and ego-centric navigation through driving scenes on urban roads and freeways, with evaluation and discussion of our pilot experiments on these datasets.

**In-Cabin Perception: Driver State Monitoring and Driving Impairment**
This case study investigates whether short segments of driver speech, captured passively inside the vehicle, contain detectable signatures of alcohol-related driving impairment, and how such information could contribute to multimodal driver state monitoring within the L-LIO framework. As an initial case study, we formulate this problem as a binary classification task, where the goal is to predict whether a given speech segment corresponds to a sober baseline condition or a condition exceeding the legal breath alcohol concentration (BrAC) threshold for driving (0.08%).[1]

Importantly, this task is framed as a proxy for driver impairment assessment, not as a definitive intoxication detector. Speech-based predictions are evaluated with the broader objective of understanding how audio cues may complement visual, behavioral, and vehicle-based signals in intelligent safety systems.

**Data Collection and Experimental Protocol** Audio data was drawn from a controlled driving-impairment study involving licensed drivers who participated in a multi-hour experimental session incorporating breath alcohol measurements, driving simulation, and multimodal sensing. Breath alcohol concentration (BrAC) was used as a noninvasive proxy for blood alcohol concentration (BAC), with a threshold of 0.08% BrAC corresponding to the legal intoxication limit for driving in the United States. Participants completed a single-blind alcohol administration protocol and consumed an individually calibrated alcohol dose targeting 0.08% BrAC based on sex and body weight. Dosing was calculated individually based on sex and body weight and mixed using 95% denatured alcohol diluted to a beverage containing 20% alcohol and 80% non-caffeinated soda. As part of the larger parent study, participants completed multiple tasks over approximately six hours, including repeated breathalyzer measurements, four driving-simulation sessions, and multimodal data collection (audio, video from four RGB cameras, thermal imaging, and simulator-derived driving behavior).

Participants were included in the present analysis if they reached an average BrAC of at least 0.08% in the first post-alcohol consumption assessment and produced valid audio in both sober and post-alcohol sessions. A total of 18 participants were included in the analysis at hand. Speech was recorded using in-cabin microphones during scripted verbal tasks designed to resemble common vehicle–driver interactions (e.g., short commands and naturalistic phrases, with phonetic and articulatory challenges; see Table 1). Each participant contributed recordings in a sober baseline condition and subsequent post-alcohol consumption conditions in which BrAC exceeded the legal driving limit of 0.08%. To mitigate potential rehearsal effects, participants read the phrases out loud once before recording the baseline condition. For the present analysis, we focus exclusively on audio from the first two speech recordings (the pre-alcohol baseline and the first post-alcohol recording, collected approximately 30 minutes after alcohol consumption, prior to the first driving simulation). This design reflects a realistic operational setting in which speech cues may precede observable changes in driving behavior. All participants provided informed consent in accordance with institutional review board approval.

***Table 1.***
***Set of phrases recorded in Case Study 1***

| |
|---|
| Hey Lexi |
| Jessica was blissfully unaware that everyone knew that her so-called designer wardrobe was indubitably homemade |
| Nuclear proliferation can't be tolerated |
| She is suffering from passive aggressive disorder |
| Branches of plum |
| This is a comfortable cushion and it is very expensive |
| There are sixty six books on the rack |
| The good public prosecutor is from Wisconsin |
| Hey Lexi set my cruise control at 65 miles per hour |
| Hey Lexi can you tell me how tall is the Eiffel tower |
| Hey Lexi take me to Starbucks on East Coast Road |
| What does the weather look like for today and tomorrow |

[1] It should be noted that BrAC level is not synonymous with impairment. Individual differences mean some people are observably impaired below the legal threshold, while others show few signs even at higher BrACs. Consequently, it is vital to distinguish between behavioral signals of impairment and the measured BrAC, as speech-based cues may reflect impairment differently across individuals. As such, while our study uses BrAC for training and model evaluation, it should be recognized that BrAC is not a direct proxy for actual impairment. These results should therefore not be overgeneralized, and caution is warranted when interpreting model performance strictly in terms of legal threshold.

**Classification and Evaluation Protocol** The task was formulated as a binary classification problem, in which each short speech segment was labeled according to whether the speaker's BrAC was sober or >= 0.08% BrAC. Speech representations were used as input to lightweight classifiers (Random Forests and linear Support Vector Machines), selected to reduce overfitting given the limited dataset size. We adopted a factorial experimental design: Specifically, we varied (1) temporal processing strategy (windowed vs. non-windowed speech segments), (2) use of subject-specific baseline subtraction (applied vs. not applied), (3) classifier type (Random Forest vs. SVM), and (4) acoustic representation (MFCCs, eGeMAPS, wav2vec2, WavLM) [2-5, 60-63].

All configurations were evaluated using a leave-one-subject-out cross-validation protocol, ensuring that all speech from a given driver was held out during testing to assess subject-independent generalization. In selected configurations, a subject-specific sober baseline embedding was computed and subtracted from test samples to simulate personalized reference calibration. Performance was assessed using accuracy and area under the ROC curve (AUC), with predictions aggregated at the utterance level.

**Speech Processing and Preprocessing** Full-session audio recordings were automatically transcribed with word-level timestamps using Faster-Whisper,[2] allowing phrase-level segmentation aligned to a known set of scripted prompts. Individual utterances were extracted via FFmpeg using these timestamps to produce short speech clips suitable for downstream analysis. All audio was standardized to a common sampling rate and converted to a single channel. To examine the effect of temporal granularity, we evaluated both utterance-level processing and short-time windowed processing, where speech clips were subdivided into overlapping 1-second segments.

**Acoustic Feature Representations** We evaluated four classes of speech representations spanning traditional handcrafted features to modern self-supervised embeddings, chosen to reflect increasing representational capacity and robustness to noise and speaker variability. While all representations start from raw audio, they differ in what variability they preserve vs. suppress, how context is modeled, and what kinds of invariances are encouraged.

*MFCC and Spectral Features* As a classical baseline, we extracted Mel-frequency cepstral coefficients (MFCCs) along with their first- and second-order temporal derivatives to capture short-term spectral dynamics associated with articulation and vocal tract configuration. MFCCs are computed by applying a short-time Fourier transform to the audio signal, warping the frequency axis onto the Mel scale (which approximates human auditory perception), taking the logarithm of spectral energies, and applying a discrete cosine transform to produce a compact representation. These features were complemented by additional descriptors, including spectral centroid, bandwidth, and contrast. Temporal statistics were computed over each segment to produce fixed-dimensional representations. This feature set captures low-level acoustic structure but relies on hand-engineered assumptions about which aspects of speech are informative and does not explicitly model long-range temporal context.

*eGeMAPS Voice-Quality Features* To capture interpretable paralinguistic cues, we extracted features from the extended Geneva Minimalistic Acoustic Parameter Set (eGeMAPS) [2, 3]. This representation consists of an expert-curated set of compact descriptors of pitch, energy, spectral balance, jitter, shimmer, and related voice-quality measures that have been widely used in affective computing and clinical speech analysis. Unlike MFCCs, which emphasize spectral shape, eGeMAPS targets higher-level acoustic correlates of physiological and prosodic characteristics that may reflect neuromotor degradation. Because eGeMAPS does not attempt to model linguistic content or long-range temporal dependencies, it is commonly used in settings where speech style, vocal effort, or physiological state is of interest, rather than speech recognition.

*Wav2Vec2 Self-Supervised Speech Embeddings* Moving beyond handcrafted descriptors, we extracted embeddings from a large pretrained Wav2Vec2 model (*facebook/wav2vec2-large-960h-lv60*) [4]. This self-supervised architecture learns contextualized speech representations directly from raw audio through large-scale pretraining on unlabeled speech corpora. During pretraining, portions of the latent sequence are masked, and the model is trained to predict the masked representations using a contrastive objective, encouraging the model to support robust speech recognition. For each utterance, the sequence of hidden states produced by the model was temporally pooled to yield a fixed-length embedding capturing phonetic structure, temporal dynamics, and speaker-state information. These embeddings encode higher-level information than traditional features and do not require manual feature design.

[2] https://github.com/SYSTRAN/faster-whisper

***WavLM Self-Supervised Speech Embeddings*** We additionally evaluated embeddings from WavLM (*microsoft/wavlm-large*) [5], a more recent foundation speech model designed to be robust to background noise, speaker variability, and short utterances—conditions characteristic of in-vehicle audio. WavLM builds on the wav2vec-style architecture but introduces enhanced pretraining objectives that explicitly model noisy, overlapping, and multi-speaker speech scenarios. As with Wav2Vec2, embeddings were obtained by mean-pooling contextualized hidden states across time, producing a compact representation intended to capture subtle impairment-related changes in speech production. Compared to earlier self-supervised speech models, WavLM is designed to balance phonetic content modeling with sensitivity to speaker-related variation, making it suitable for downstream tasks that extend beyond speech recognition.

**Evaluation Procedure** We used leave-one-subject-out (LOSO) cross-validation to ensure that no data from the test speaker appeared in the training set. For each fold, all clips from one subject were held out for testing, and the models were trained on the remaining subjects. For each fold, all features (in the training windows) were z-normalized using a StandardScaler fit, and PCA was applied to reduce feature dimensionality while retaining most variance. The number of components was capped at 50. In the baseline-subtraction condition, a subject-specific sober baseline vector was computed from the held-out subject's sober windows and applied only to their test features. Since this normalization does not use any training data or affect fitted models, it does not constitute data leakage and mirrors how a real system might use a stored sober reference sample for each driver. Window-level probabilities were aggregated to clip-level predictions using the median (following standard practice in paralinguistic sequence classification). Accuracy, ROC AUC, and per-subject metrics were computed at the clip level. Final predicted labels were obtained with a threshold of 0.5.

Whereas this first case study evaluates audio as a diagnostic signal under controlled conditions, the following study uses formative, system-in-the-loop analysis to examine how spoken language reshapes perception-planning interfaces, prioritizing representational insight.

**Bridging Inside and Outside the Vehicle: Audio for Grounding Language to External Scene**
In our second case study, we examine how natural language directives can be used to connect audio instruction from inside the vehicle to how the vehicle plans motion in its external environment. This investigation focuses on how Instruction-Action (IA) pairs serve as valuable and information-rich resources for robotic systems. We first consider passenger-style instruction grounded in driving scenes and trajectories. We then explore how Global Positioning System (GPS) references and natural language processing support the curation of volumes of IA pairs and how this information could contribute to the L-LIO framework. Across both settings, audio instructions provide additional contexts that link what is said in the vehicle to how motion is planned and executed.

**Passenger Natural Language Instructions for Scene-Conditioned Planning** We examine the structure of passenger-style natural language instructions when aligned with real-world driving scenes. Our study is based on clips from the nuScenes dataset, which provides ego-centric camera views and corresponding vehicle trajectories. The dataset scenes were manually annotated using the "taxi test" heuristic: if you were a passenger in a taxi during this clip, what instruction would you give to produce this motion? This process yielded instructions such as "turn at the stop sign," forming scene-language-trajectory triads that expose how spoken instructions reference spatial context, dynamic agents, and intent—the basis for our augmented doScenes dataset [14].

Connecting in-cabin language with vehicle decision-making, the instructions were injected into an existing VLA planning framework [64]. Across the annotated scenes, passenger instructions varied in length, specificity, and referential style. Some instructions were short, while others were descriptive and grounded in dynamic elements such as pedestrians or intersections, which highlight the rich semantic and contextual cues that are not directly observable from vision alone. Thus, audio-aligned instruction data is essential to understand how humans naturally communicate intent in driving scenarios.

While passenger speech provides a natural interface for expressing intent, reliance on human instruction is not scalable in many deployment settings, motivating the use of structured audio sources such as GPS navigation commands. These commands serve as a complementary mechanism for generating large volumes of IA pairs and support future scene-responsive planning systems without requiring continuous passenger input.

**Information Analysis and the Importance of Instructional Commands from GPS systems** We analyzed linguistic characteristics in readily-available information from navigational systems. This provided insight into how natural language instructions can vary slightly across navigational applications, offering sufficient or deficient support for particular VLA tasks. Based on different referential cues, we categorized observed statements from our initial pilot dataset into a set of classes (e.g., static objects, road names, etc.), offering an accessible analysis method to use structured natural language instructions from navigational systems. Furthermore, because of their underlying programmatic structure, navigational applications naturally create similarly structured outputs, motivating our survey of the range of structured outputs.

Small changes in GPS references can have different effects on model performance and perception capabilities, leading to changes in how instructions are executed and affecting the complexity of the decision-making. The analysis of [65] demonstrates such data and a method by which GPS references can increase the number of IA pairs curated by automating the annotation workflow through this audio integration. Such IA pairs can be used as a source of data for model pretraining, improving contextual data and environmental awareness.

**Pilot Dataset** We formed a pilot dataset of various GPS navigational systems, including manual class annotations of five driven routes comprising residential, rural, and commercial areas, city streets, and freeways in California. We evaluated the data by identifying patterns in the verbal instructions, summarizing key differences in verbalized navigation having instances of multi-attribute instructions. Leveraging different navigational applications and comparing Apple Maps, Google Maps, and Waze Maps helped with structure analysis and possible utility for VLA models or in cases where verbalized instructions with a particular style of referentiality may be most useful.

**ADVLAT-Engine** The ADVLAT (Autonomous Driving Vision-Language-Action Triad) Engine framework [65] is designed to increase the curation of IA pairs by automating the data annotation process, eliminating the need for manual annotations. The ADVLAT-Engine processes natural language instructions in the form of spoken commands provided by a navigational system or passenger. Following audio collection, a transcription model is applied to convert the spoken instruction into text, with corresponding timestamps for each command.

In addition to the audio, the framework incorporates visual perception from external sources by obtaining in-car video recordings of a user following a set of verbal navigational commands, serving as valuable information for analyzing potential road behaviors and road conditions. The timestamped audio instructions leverage the broader multi-input framework by grounding spoken instructions to the external scene, along with recording the trajectories (i.e. GPS positions). Additionally, by incorporating off-the-shelf tracking applications with an external camera, the ADVLAT-Engine can synchronize the visual perception and trajectory data, allowing the engine to assemble the vehicle perception over its traveling distance over time, while tagging timestamps for audio events. Together with the passenger-instruction case, the ADVLAT-Engine demonstrates how listening to either human passengers or GPS systems can be leveraged to curate IA datasets and to support scene-responsive motion planning within the L-LIO framework.

**Natural Language Instructions for Scene-Responsive Motion Planning** Natural language instructions act as a bridge between human interaction and robotic perception for motion planning, translating the intent of passengers into context-aware actions. Natural language instructions carry valuable information such as context cues, spatial, and semantic details, which can be used for scene-responsive motion planning, helping autonomous systems interpret their environments dynamically. By incorporating audio as a modality, we can combine language with perception, letting an autonomous system make a grounded adjustment based on the verbalized instructions and the actual scene. The ability to expand the availability of training data at minimal cost has the potential to significantly progress the robustness, safety, and interactivity of autonomous systems.

**External Scene Understanding: Audio for Event and Reference Recognition**

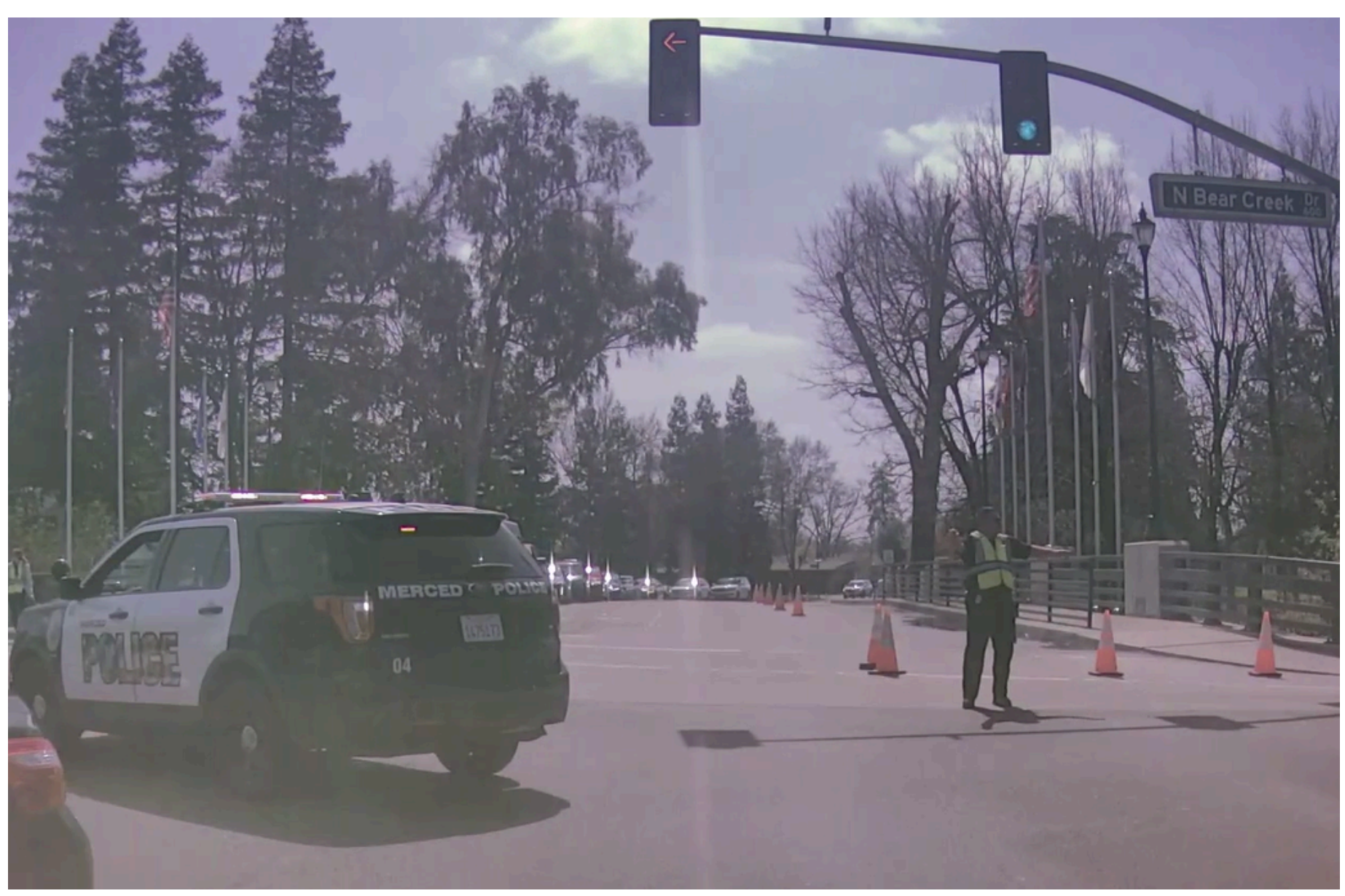

***Figure 2. Within the L-LIO framework, autonomous systems approaching the depicted scene can use audio to link temporal cues (such as sirens and whistles) to identify situation abnormality and associate attention-reference of guiding human agents and their associated gestures to the ego vehicle.***

While the previous case study focused on explicit linguistic intent communicated through speech, many external safety interactions rely on non-verbal but directive signals (i.e., gestures, whistles, horns, and sirens), whose meaning emerges only through multimodal context. Motivated by the edge case failure of autonomous vehicles to safely navigate dynamic scenes where human guidance is necessary for planning (e.g., an intersection blocked by first responder activity, with a traffic director guiding contraflow, as depicted in Figure 2), we collected and analyzed data using state-of-the-art vision-language models to understand the intentions of the guiding agents. Our pilot evidence suggests that vision-only vision–language interpretation of dynamic traffic gestures is not yet sufficiently reliable for safety-critical scene understanding and decision-making. We constructed two datasets, Acted Traffic Gestures (ATG) and Instructive Traffic Gestures In-the-Wild (ITGI), and evaluated state-of-the-art VLMs on multiple understanding tasks in a zero-shot setting, reflecting realistic deployment constraints where task-specific gesture training data may be unavailable or limited [66]. Gesture understanding was assessed using three complementary evaluation formulations. First, caption similarity measured how closely model-generated textual descriptions matched reference gesture descriptions, capturing semantic alignment. Second, gesture classification accuracy evaluated whether models could map observed gestures to a predefined set of traffic-relevant commands. Third, pose-sequence similarity compared inferred gesture motion patterns against reference trajectories, assessing whether models captured the temporal structure of the gesture rather than static appearance alone. This multi-pronged evaluation was intended to probe both semantic correctness and temporal consistency, which are critical for safe interpretation.

**RESULTS**

**In-Cabin Perception: Driver State Monitoring and Driving Impairment**

Across all configurations, speech-only models demonstrated limited but consistent sensitivity to alcohol-related impairment, with performance strongly dependent on the choice of acoustic representation. Overall performance was modest, reflecting the difficulty of subject-independent impairment detection from short speech segments, but trends were stable across modeling choices and evaluation settings.

**Impairment Classification Performance** Representation choice emerged as the dominant factor influencing classification performance. Representations that preserve paralinguistic and speaker-dependent variation outperform those that emphasize speaker-invariant, content-focused encoding for impairment detection (see Table 2 and Fig.1). Impairment detection depends on subtle, speaker-dependent, paralinguistic cues (prosody, articulation stability, voice quality), and representations optimized for content or speaker invariance (often for tasks such as ASR) actively work against this.

WavLM embeddings consistently achieved the strongest performance, yielding the highest area under the ROC curve (AUC ≈ 0.65) under leave-one-subject-out evaluation. This suggests that representations explicitly designed to handle noise, speaker variability, and short utterances are well-suited to in-vehicle speech analysis. Handcrafted acoustic features (MFCCs and eGeMAPS) performed below WavLM but above chance, achieving moderate discrimination (MFCC: AUC ≈ 0.57; eGeMAPS: AUC ≈ 0.59). These results indicate that classical spectral and voice-quality features do contain impairment-relevant information, particularly when evaluated across a sufficiently diverse subject pool.

In contrast, wav2vec2 embeddings showed comparatively weaker performance (AUC ≈ 0.54), and were outperformed by both WavLM and the handcrafted feature sets. This suggests that not all pretrained speech representations are equally sensitive to the paralinguistic and physiological cues associated with alcohol-related impairment. While self-supervised embeddings can capture rich speech structure, their utility for impairment detection depends on whether paralinguistic and speaker-dependent cues are preserved rather than normalized away. Although both wav2vec2 and WavLM rely on self-supervised pretraining, they differ substantially in the inductive biases imposed by their objectives. Wav2vec2 emphasizes speaker-invariant, content-focused representations optimized for phonetic discrimination, which can suppress paralinguistic variation relevant to impairment. WavLM, however, incorporates noise-robust and multi-speaker pretraining objectives that preserve speaker-dependent and prosodic cues, making it better suited for capturing subtle changes in speech production associated with degraded driver state.

Classifier choice (Random Forest vs. linear SVM), temporal windowing strategy, and subject baseline subtraction had relatively minor effects on overall performance compared to the impact of representation selection.

These findings suggest that impairment-related cues are present in driver speech, but the signal is highly variable, especially when evaluated under subject-independent conditions representative of real vehicle deployment in a small dataset.

***Table 2.***
***Performance comparison of Case Study 1: best configuration across embeddings***

| Embedding | Classifier | Baseline | Window | Accuracy | AUC |
|---|---|---|---|---|---|
| wavlm_large | RF | baseline | nowindow | 0.611 | 0.654 |
| egemaps | RF | nobaseline | window | 0.565 | 0.589 |
| mfcc | RF | baseline | window | 0.542 | 0.568 |
| wav2vec2_large | RF | baseline | nowindow | 0.551 | 0.547 |

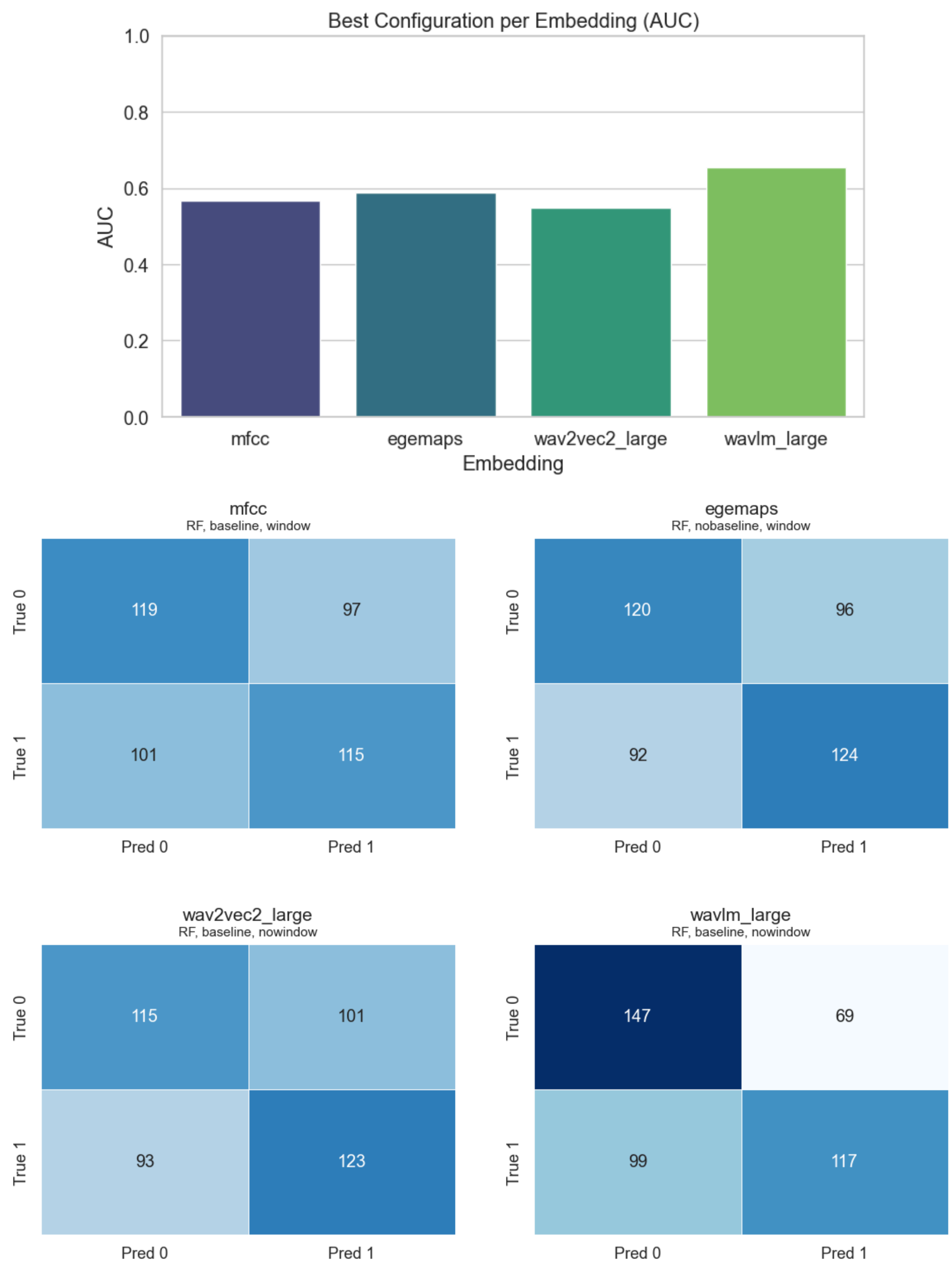

***Figure 2. Overall performance across representation types (top) and clip-level confusion matrices (bottom) showing classification results for the best configuration of each representation.***

**Interspeaker Variability** A dominant observation across all experiments was substantial variability between drivers (see Fig.2). Some participants exhibited clear separability between sober and intoxicated speech, while others showed little discernible acoustic change. This variability significantly constrained population-level performance and underscores a key limitation of audio-only impairment detection. Such interspeaker differences are consistent with prior findings in speech and intoxication research and highlight the difficulty of relying on any single modality for robust driver state assessment. Notably, representations that preserved speaker- and prosody-related variability—such as WavLM and the handcrafted acoustic features—appeared better able to exploit this diversity than representations emphasizing speaker invariance. These findings reinforce the challenge of modeling impairment as a universal acoustic signature.

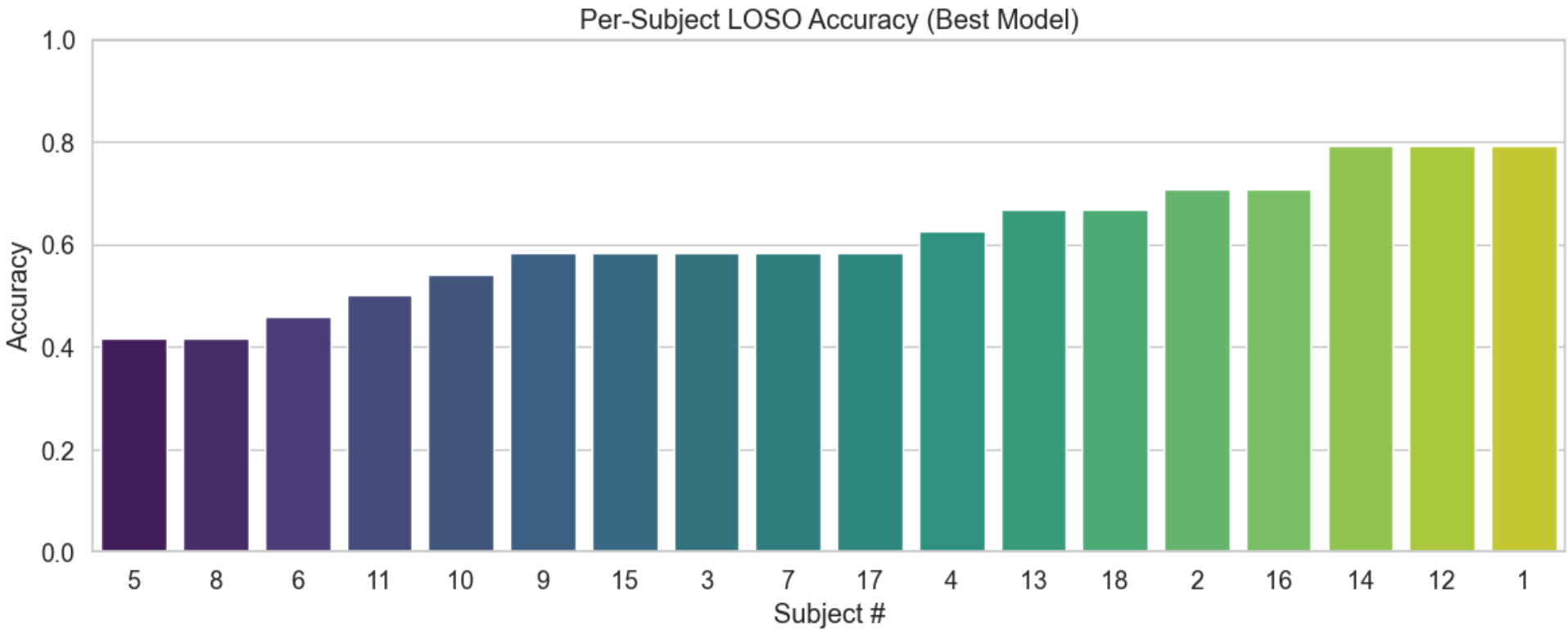

***Figure 3. Interspeaker variability in clip-level LOSO accuracy for each of the 18 subjects under the best-performing model (WavLM, Random Forests, no-window, baseline). The wide range of per-subject accuracy indicates that speaker-specific differences dominate, creating large variability across LOSO folds and making generalization challenging.***

**Implications for Multimodal Driver Monitoring** Taken together, these results indicate that audio should not be treated as a standalone detector of impairment, but rather as a complementary signal within a broader multimodal driver monitoring framework. Speech-based indicators may serve as early or soft cues of degraded driver state, prompting increased attention from other sensing modalities such as vision, vehicle dynamics, or physiological signals. While speech alone provides only moderate discrimination, at least on this dataset, it offers access to internal driver state through channels that are largely orthogonal to vision, vehicle dynamics, and physiological sensing.

Within the proposed L-LIO framework, this case study illustrates that in-cabin speech carries state-related information but also exposes the limits of interpreting such signals in isolation. Rather than positioning audio solely as a detector of internal driver condition, these findings motivate a broader view of listening inside the vehicle—one that emphasizes its role in conveying intent, reference, and context. In the following case study, we shift focus from driver state inference to how spoken language inside the cabin can be grounded in the external driving scene to support scene understanding and decision-making.

**Bridging Inside and Outside the Vehicle: Audio for Grounding Language to External Scene**
Audio serves as an important modality for grounding language to the external scene, enabling an autonomous system to enhance spatial grounding and trajectory approximation in a dynamic environment. Incorporating audio cues helps align verbal instruction with the vehicle's trajectory. Additionally, audio cues extend beyond an autonomous system's field of vision, letting it detect unusual events before they are in its field of view.

**Passenger Natural Language Instructions for Scene-Conditioned Planning** We performed a pilot integration of passenger-style spoken directives with an instruction-conditioned planner to understand how in-cabin speech can influence scene-responsive motion planning. As the purpose of this paper is to motivate the collection of audio for the generation of instruction-action pairs, we direct the reader to the full results of the language-conditioned trajectory planning exploration in [64], summarizing here that a visual-only trajectory planning baseline, using the OpenEMMA VLA, achieves an ADE of 2.879 (discounting significant model-failure outliers of this baseline), and that the instruction-provided variant achieves a lower ADE of 2.732. These results are further exaggerated when the baseline model failures are included in the statistics, showing a strong advantage for instruction-conditioned trajectory prediction.

The key takeaway is motivational rather than a benchmarking result: vehicle behavior is sensitive to what the passenger says, how it is phrased, and how it is transcribed from audio. This motivates collecting time-aligned cabin audio as an intent signal that can be grounded against the visual scene and safety constraints. Figure 4 provides an illustrative case, motivating L-LIO collection and time-alignment of in-cabin audio for instruction grounding.

**Scene-642**

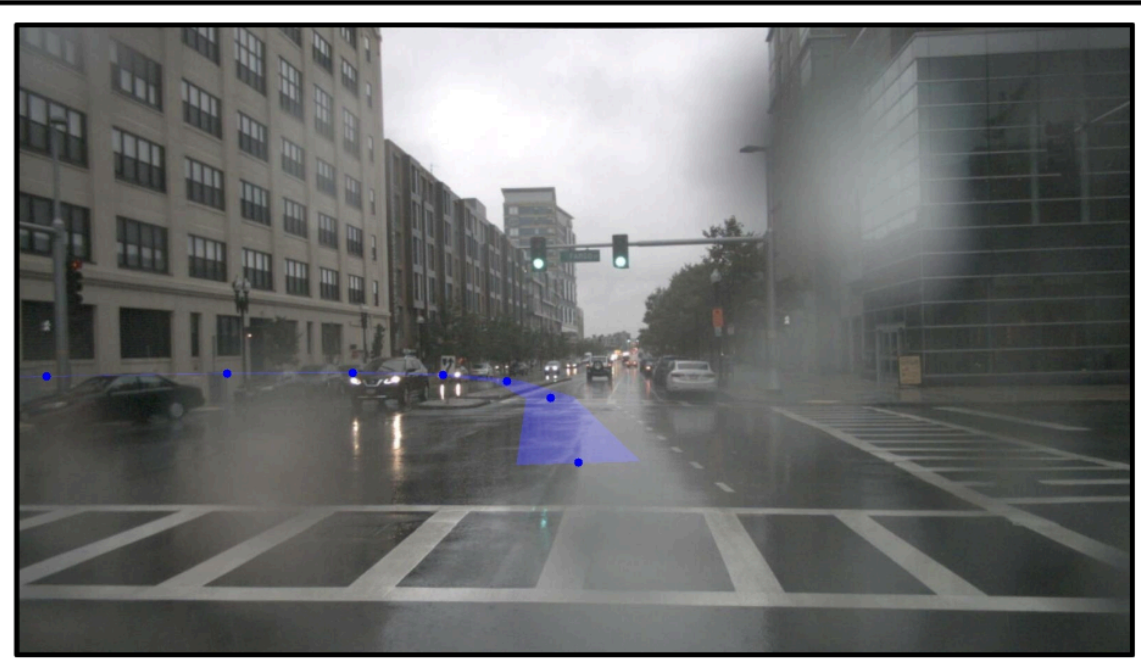

*No Instruction*

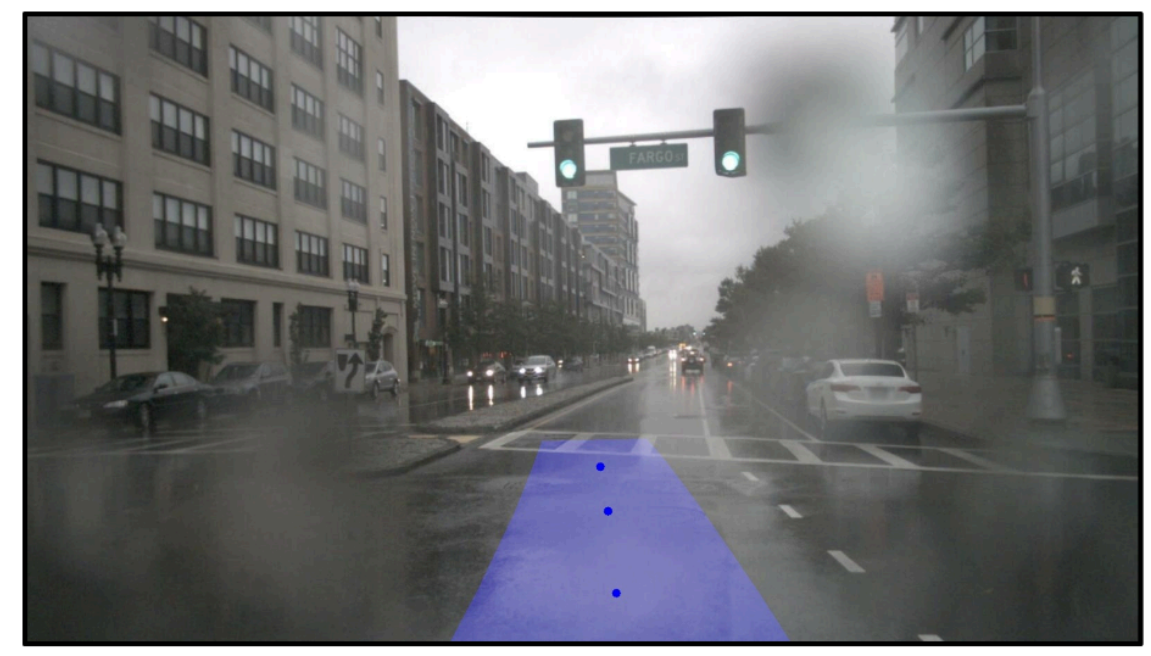

*"Go straight when the stop light turns green"*

***Figure 4. Qualitative comparison of trajectory predictions with and without natural language instruction.***

Lastly, we observe that the length and reference type of a human-generated instruction influenced the trajectory behavior. The average prompt (9-12 words) yields the greatest improvement relative to the same scene without instruction. When comparing the reference type in the instruction, dynamically-grounded instructions that referred to other agents, such as vehicles or pedestrians, achieved the lowest ADE differential, while non-referential or purely static instructions showed weaker gains.

**GPS Navigation Commands as a Scalable IA Resource** Using the ADVLAT-Engine, these GPS instructions can be automatically aligned with video and trajectory logs to form vision–language–action triads, without manual annotation. Overall, passenger speech demonstrates how listening can directly shape scene-responsive motion planning, while GPS navigation audio offers a scalable way to curate large IA datasets and provide instruction-like guidance in real time. Categorical analysis of available GPS commands is available at [65].

**External Scene Understanding: Audio for Event and Reference Recognition**

Bossen et al. report that, across their pilot datasets and evaluation methods, current VLMs struggle to robustly interpret traffic gestures [66]. In their experiments, average sentence similarity remains below 0.59, while gesture classification F1 reaches only 0.14–0.39, substantially below an expert baseline of 0.70, suggesting that none of the tested models are currently accurate and robust enough to be trustworthy in safety-critical usage. Across their pilot datasets and evaluation strategies, current VLMs exhibit low reliability for interpreting dynamic traffic gestures, with average sentence similarity below ~0.59 and gesture classification F1 scores well below an expert baseline.

The vision-only approach to outside agent gesture and intent recognition left gaps in understanding, especially regarding *when* the human agent is referring to the ego vehicle, which is typically communicated to human drivers through a pairing of an audio whistle or spoken command while the instructor visually faces the instructed vehicle. Our identification of errors and inconsistencies across multiple evaluation formulations motivates caution when using VLM outputs as direct behavioral triggers; current VLM outputs are not robust enough to be trusted for safety-critical interpretation of traffic gestures, and may benefit from further disambiguation, both in instruction intent and referentiality, facilitated through inclusion of audio within the perception framework.

We treat this result not as a limitation of gesture sensing alone, but as a broader observation: the intent behind external human signals is often underdetermined by vision, especially under occlusion, low illumination, cluttered scenes, and cultural variability in gestures. Accordingly, while our current gesture-focused prototype contains no external audio, we include the findings of [66] as motivation for L-LIO, suggesting external audio may act as a disambiguation and redundancy channel for safety in addition to its role as a primary control input.

From these limitations, we frame audio as an enabling modality for gesture understanding in real traffic settings. Specifically, we argue that audio can (a) provide disambiguating context when visual evidence is underspecified (e.g., similar arm motions used for different intents), and (b) carry high-precision temporal directive semantics that may co-occur with gestures in formally-directed scenes (e.g., traffic conductors), including whistles, shouted instructions, sirens, and horns. This aligns with our finding that gesture interpretation often requires comprehensive scene understanding and that directed scenarios often involve an authority figure whose role and direction of attention (signaled by both visual and audio cues) shape the interaction. More broadly, external acoustic events (sirens/horns/shouts) can signal urgency and intent even when the source is occluded or outside the camera's field of view, potentially mitigating the "caption drift" and inconsistent interpretations reflected in low similarity and F1 scores.

## DISCUSSION

Across the three case studies, audio emerges not as a single capability but as a family of safety-relevant signals whose value depends on how it is integrated into the perception–decision loop. In-cabin speech analysis, natural language instruction grounding, and external acoustic event interpretation each expose different strengths and limitations of listening as a sensing modality. Taken together, these cases illustrate that audio is most effective when used to shape uncertainty, align intent, and provide redundancy in safety-critical situations, and that through continued research and data collection, future models may leverage this modality to directly command vehicle behavior. This perspective reframes listening within the proposed L-LIO framework as a mechanism for robustness and risk mitigation—complementing visual perception and planning with signals that are often temporally urgent, semantically rich, and difficult or impossible to infer from vision alone.

Our analysis of driver speech illustrates audio's role as a diagnostic signal for internal driver state, rather than an explicit communication channel. Our findings show that short speech signals contain detectable but limited markers correlated with alcohol-related impairment, particularly when representations preserve paralinguistic characteristics. However, performance under subject-independent evaluation reveals substantial inter-speaker variability, highlighting the difficulty of inferring impairment reliably from short, naturalistic speech segments alone. These results reinforce an important design principle for L-LIO systems: diagnostic audio should be interpreted as soft, probabilistic evidence that complements visual, behavioral, and vehicle-based signals. In this role, listening inside the vehicle functions as an early warning and contextual enrichment mechanism, capable of prompting heightened monitoring or conservative intervention without asserting unwarranted certainty.

Our integration of passenger-style natural language into multimodal planning VLA OpenEMMA illustrates how listening from inside the vehicle can directly influence decision-making outside the vehicle. In our experiments, instruction-conditioned planning was able to constrain unsafe actions and suppress rare but severe planning failures, highlighting the role of language as a behavioral guide more so than a trajectory optimizer. The results expose important challenges associated with language-conditioned planning in real-world settings, namely the fact that passenger instructions are rarely clean or consistent. Individuals inside an autonomous vehicle may speak quickly, repeat themselves, shorten phrases, or use imprecise references that may be clear to humans but ambiguous to models. This inherently adds another layer of noise with spoken interfaces. Our observations suggest that even small variations in wording can influence planning behavior, and in practice, this raises the risk that a well-intentioned passenger could unintentionally affect vehicle behavior in undesirable ways if linguistic variability is not handled robustly. Overall, these observations highlight the need for language-conditioned planners to prioritize intent understanding and grounding in the visible scene, rather than reacting directly to surface-level phrasing. Passenger language should therefore be treated as an untrusted, high-level input, rather than a direct control signal. While this work assumes cooperative passenger intent, systems will require safeguards to ensure that natural language interfaces cannot induce unsafe actions, whether due to ambiguity, error, or misuse. This especially motivates the continued collection of such data in instruction-action pairs, such as through the proposed ADVLAT-Engine, so that models can continue to be trained, fine-tuned, and evaluated over a wide range of driving scenarios and instructive speakers.

Our presented studies also broaden the role of audio by highlighting its value as a disambiguation and redundancy channel in safety-critical interactions with humans outside the vehicle. Prior analyses of vision-language models demonstrate that gesture interpretation in real traffic environments remains fragile, particularly under occlusion, clutter, and referential ambiguity. In many such scenarios, human intent is not communicated through gesture alone, but through coordinated multimodal signals that pair motion with acoustic cues such as whistles, shouted directives, horns, or sirens. From a safety assurance perspective, listening outside the vehicle enables the system to detect urgency, authority, and intent even when visual evidence is incomplete or unreliable. Importantly, the expected benefit of external audio is not more aggressive or complex maneuvering, but increased robustness through conservative fallback behaviors—slowing, yielding, or requesting driver takeover—when uncertainty is high. These outcomes support the idea that listening outside the vehicle is not optional for robust interaction understanding; instead, it is a promising route to improving reliability for cooperative driving scenarios where humans communicate through a mixture of gesture and sound, and where vision-only foundation models currently underperform for safe motion planning. From the shortcomings of VLMs in these scenarios, we suggest two applications of audio towards improvement of external agent instruction understanding: (1) Acoustic Event Detection (AED), such as detection of safety-relevant events such as sirens, horns, whistles, and urgent shouts, producing time-stamped event labels with confidence scores, and (2) Optional Speech-to-Text (ASR) for directive speech, such that when speech is present (e.g., “stop,” “slow,” “go,” “back up”), audio can be converted to text and map to a small set of standardized intent tokens. These outputs can be fused with visual cues (e.g., pedestrian pose/gesture, scene context) for stronger scene understanding and safe decision-making.

Viewed together, these case studies suggest that primary contributions of audio within L-LIO systems are not accuracy in isolation, but rather resilience under uncertainty and complementary modes of information in unsafe scenarios that demand human guidance. Diagnostic speech cues, natural language instructions, and external acoustic events each introduce their own forms of noise, ambiguity, and variability; yet across all three settings, audio provides access to safety-relevant information that is orthogonal to vision and vehicle dynamics. The common design lesson is that listening should be used to modulate confidence, trigger heightened caution, and reduce reliance on brittle single-modality interpretations, rather than to replace existing perception pipelines. By treating audio as a first-class but carefully mediated modality, the L-LIO framework supports even more conservative, interpretable, and robust decision-making in complex real-world driving environments where safety depends as much on managing uncertainty as on perceiving the scene correctly.

## LIMITATIONS

This work presents pilot evidence supporting audio as a first-class modality within the proposed L-LIO framework, but several limitations remain.

First, all case studies are evaluated at a limited scale. In particular, the driver speech impairment analysis involves a small number of participants and short, scripted utterances, which limits population-level generalization. Performance under leave-one-subject-out evaluation highlights substantial inter-speaker variability, indicating that speech alone is insufficient for reliable driver state assessment and should be treated as a complementary signal rather than a standalone detector.

Second, audio-based interfaces are sensitive to noise, phrasing variability, and transcription errors. In the instruction-grounding case study, small differences in wording or ASR output influenced downstream behavior, underscoring the need for intent extraction, uncertainty handling, and safety arbitration before spoken language can be used reliably in planning systems.

Third, planning-related analyses are conducted in open-loop or pilot integration settings. We do not evaluate long-horizon closed-loop behavior or recovery from compounding error, and external audio integration for gesture disambiguation remains conceptual rather than fully implemented, with the intention that exposing current limitations and gaps of existing vision-centric methods may lead to inclusion of large-scale studies leveraging audio to overcome performance barriers.

Finally, continuous audio sensing introduces privacy and deployment considerations that are not addressed in this research and must be carefully managed in real-world systems.

## CONCLUSIONS

This paper introduces the looking-and-listening inside-and-outside (L-LIO) framework, extending the traditional LILO paradigm by elevating audio to a first-class modality for intelligent vehicle safety. Through three pilot case studies, we demonstrate that listening, both inside and outside the vehicle, provides safety-relevant information that is difficult or impossible to infer from vision alone, including cues about driver state, human intent, and urgent environmental events.

Our findings show that audio is most valuable when used as a complementary signal rather than a standalone solution. In-cabin speech contains detectable markers of degraded driver state but exhibits high inter-speaker variability, motivating multimodal fusion. Spoken instructions offer a natural interface for conveying intent but require careful grounding and safety arbitration to avoid unintended behavioral effects. External audio cues such as sirens, horns, and shouted directives provide critical context and temporal urgency that can reduce uncertainty in complex or visually ambiguous scenes.

Taken together, these results support the central premise of L-LIO: safe and interactive vehicle intelligence and autonomy requires both looking and listening. As autonomous and semi-autonomous systems increasingly rely on language-centric AI models, incorporating audio into vehicle sensing and datasets will be essential for robust, human-aligned decision-making. This work motivates future efforts toward large-scale multimodal datasets, closed-loop evaluation, and principled fusion strategies that treat audio as an integral component of safety-critical vehicle intelligence.

**ACKNOWLEDGMENTS**
Human subjects studies described in this report were approved by IRB #806305 of the Office of IRB Administration (OIA) of the University of California, San Diego. A study described in this report was partially supported by the Toyota Collaborative Safety Research Center.